%% file: 00_main.tex
\documentclass[sigconf]{acmart}

\input{commands}

\AtBeginDocument{%
  \providecommand\BibTeX{{%
    \normalfont B\kern-0.5em{\scshape i\kern-0.25em b}\kern-0.8em\TeX}}}

\setcopyright{acmcopyright}
\acmConference[SIGGRAPH '22 Conference Proceedings]{Special Interest Group on Computer Graphics and Interactive Techniques Conference Proceedings}{August 7--11, 2022}{Vancouver, BC, Canada}
\acmBooktitle{Special Interest Group on Computer Graphics and Interactive Techniques Conference Proceedings (SIGGRAPH '22 Conference Proceedings), August 7--11, 2022, Vancouver, BC, Canada}
\copyrightyear{2022}
\acmYear{2022}
\acmDOI{10.1145/3528233.3530718}
\acmPrice{15.00}
\acmISBN{978-1-4503-9337-9/22/08}

\newcommand{\datasetname}{Heritage-Recon\xspace}

\acmSubmissionID{393}

\begin{document}

\title{Neural~3D~Reconstruction~in~the~Wild}

\begin{anonsuppress}
    \author{Jiaming Sun}
    \orcid{0000-0003-4053-8510
    }
    \email{jiaming@idr.ai}
    \affiliation{
    \institution{\href{https://idr.ai}{Image Derivative Inc.}}
    \country{China}
    }
    \author{Xi Chen}
    \email{cxi@zju.edu.cn}
    \affiliation{
    \institution{Zhejiang University}
    \country{China}
    }
    \author{Qianqian Wang}
    \email{qw246@cornell.edu}
    \affiliation{
    \institution{Cornell Tech, Cornell University}
    \country{USA}
    }
    \author{Zhengqi Li}
    \email{zl548@g.cornell.edu}
    \authornote{
        Current address: Google Research.
    }
    \affiliation{
    \institution{Cornell Tech, Cornell University}
    \country{USA}
    }
    \author{Hadar Averbuch-Elor}
    \email{hadarelor@cornell.edu}
    \affiliation{
    \institution{Cornell Tech, Cornell University}
    \country{USA}
    }
    \author{Xiaowei Zhou}
    \authornote{
        Affiliated with State Key Lab of CAD\&CG, Zhejiang University.
    }
    \email{xwzhou@zju.edu.cn}
    \affiliation{
    \institution{Zhejiang University}
    \country{China}
    }
    \author{Noah Snavely}
    \email{snavely@cornell.edu}
    \affiliation{
    \institution{Cornell Tech, Cornell University}
    \country{USA}
    }

    \renewcommand\shortauthors{Sun, J. et al}
\end{anonsuppress}

\begin{abstract}
We are witnessing an explosion of neural implicit representations in computer vision and graphics. 
Their applicability has recently expanded beyond tasks such as shape generation and image-based rendering to the fundamental problem of image-based 3D reconstruction. However, existing methods typically assume constrained 3D environments with constant illumination captured by a small set of roughly uniformly distributed cameras. We introduce a new method that enables efficient and accurate surface reconstruction from Internet photo collections in the presence of varying illumination. 
To achieve this, we propose a hybrid voxel- and surface-guided sampling technique that allows for more efficient ray sampling around surfaces and leads to significant improvements in reconstruction quality. Further, we present a new benchmark and protocol for evaluating reconstruction performance on such in-the-wild scenes. We perform extensive experiments, demonstrating that our approach surpasses both classical and neural reconstruction methods on a wide variety of metrics.
Code and data will be made available at {\color{urlcolor}\urlNewWindow{https://zju3dv.github.io/neuralrecon-w}}.

\end{abstract}



\input{figures/teaser}

\maketitle

\input{sections/01_intro}

\input{sections/02_related_work}

\input{sections/03_method}

\input{sections/04_experiments}
\input{sections/05_conclusion}

\clearpage\enlargethispage{16pt}
\bibliographystyle{ACM-Reference-Format}
\bibliography{bibliography}

\end{document}

%% file: commands.tex
\usepackage{soul}
\usepackage{amsmath}

\usepackage{amsfonts}
\usepackage{mathtools}
\usepackage{bm}
\usepackage{xfrac}
\usepackage{xspace}
\usepackage[makeroom]{cancel}
\usepackage{makecell}
\usepackage{multirow}
\usepackage{tabularx}
\usepackage{subcaption}
\usepackage{microtype}

\usepackage{color}
\usepackage{xcolor}
\usepackage{colortbl}
\usepackage{nicefrac}
\usepackage{ wasysym }

\usepackage[abs]{overpic}

\usepackage[symbol,para]{footmisc}

\hypersetup{pdfnewwindow}
\newcommand{\urlNewWindow}[1]{\href[pdfnewwindow=true]{#1}{\nolinkurl{#1}}}
\definecolor{urlcolor}{HTML}{d10056}
\usepackage{hyperref}
\hypersetup{
  breaklinks   = true,
  colorlinks   = true, %
  urlcolor     = RubineRed, %
  linkcolor    = RubineRed, %
}

\usepackage{graphicx}

\renewcommand{\paragraph}[1]{\vspace{0.5em}\noindent\textbf{#1}.}

\definecolor{myyellow}{rgb}{1,1, 0.6}
\definecolor{myorange}{rgb}{1, 0.8, 0.6}
\definecolor{myred}{rgb}{0.94, 0.36, 0.1}
\definecolor{newcolor}{HTML}{3f37c9}
\definecolor{mygreen}{HTML}{005caf} %

\newcommand\rurl[1]{%
  \href{https://#1}{\nolinkurl{#1}}%
}

\newcommand{\figref}[1]{Fig.~\ref{#1}}

\newcommand{\added}[1]{{#1}}

\long\def\ignorethis#1{}

\def\figcell#1#2#3{\begin{subfigure}{#1\columnwidth}\centering\includegraphics[width=\textwidth]{#2}
\def\temp{#3}\ifx\temp\empty\else\caption*{\centering{#3}}\fi\end{subfigure}}

\def\figcellc#1#2#3{\begin{subfigure}{#1\columnwidth}\centering\includegraphics[width=\textwidth]{#2} \def\temp{#3}\ifx\temp\empty\else\caption{#3}\fi\end{subfigure}}

\def\figcellt#1#2#3#4{\begin{subfigure}{#1\columnwidth}\centering\includegraphics[width=\textwidth,#4]{#2} \def\temp{#3}\ifx\temp\empty\else\caption*{#3}\fi\end{subfigure}}

\newcommand{\figcellbox}[1]{\fcolorbox{black}{white}{#1}}

\newcommand{\figcellboxwhite}[1]{\fcolorbox{white}{white}{#1}}

\def\figcelltb#1#2#3#4{\begin{subfigure}{#1\columnwidth}\centering\figcellbox{\includegraphics[width=\textwidth,#4]{#2}} \def\temp{#3}\ifx\temp\empty\else\caption*{#3}\fi\end{subfigure}}

\def\figcelltbwhite#1#2#3#4{\begin{subfigure}{#1\columnwidth}\centering\figcellboxwhite{\includegraphics[width=\textwidth,#4]{#2}} \def\temp{#3}\ifx\temp\empty\else\caption*{#3}\fi\end{subfigure}}

\def\figcellb#1#2#3{\begin{subfigure}{#1\columnwidth}\centering\figcellbox{\includegraphics[width=\textwidth]{#2}}
\def\temp{#3}\ifx\temp\empty\else\caption*{\centering{#3}}\fi\end{subfigure}}

\makeatletter
\newcommand{\thickhline}{%
    \noalign {\ifnum 0=`}\fi \hrule height 1pt
    \futurelet \reserved@a \@xhline
}
\newcolumntype{"}{@{\hskip\tabcolsep\vrule width 1pt\hskip\tabcolsep}}
\makeatother

%% file: figures/teaser.tex
\fboxsep=0pt %
\fboxrule=0.4pt %

\begin{teaserfigure}
  \begin{center}
    \captionsetup[sub]{labelformat=parens}
    \includegraphics[width=.93\textwidth]{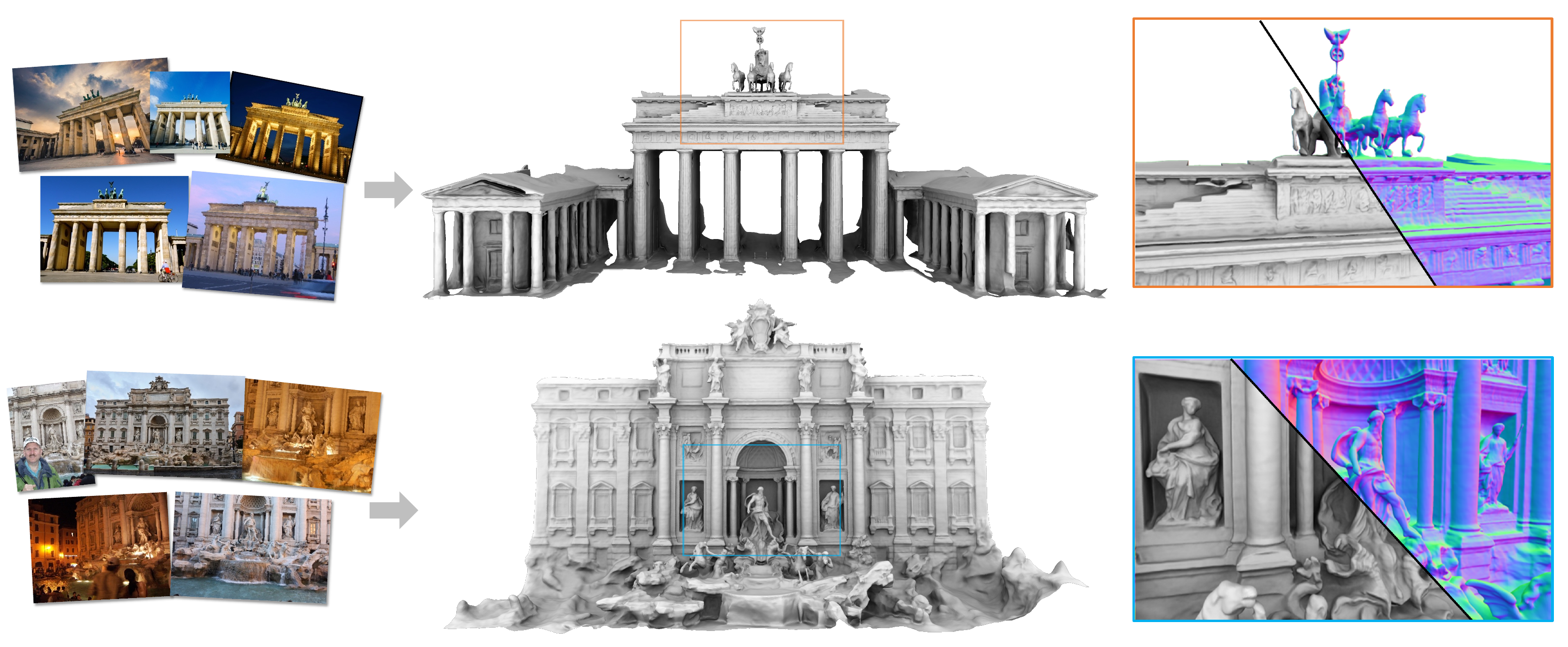}
    \caption{
      Neural 3D reconstruction in the wild.
      \normalfont{
      Given a large number of Internet photos
      capturing popular tourist attractions (left), 
    our approach learns to produce high quality 3D surface reconstruction, 
    efficiently modeling entire 3D scenes with a novel hybrid neural implicit representation (right).
    Colors indicate surface normals.
    Please zoom in to see details in the geometry. 
    {\color{gray} Photos by Flickr users rickz, Infinite Ache, jon collier, Modes Rodríguez, Ashwin Kumar, Richard Cyganiak, David Lebech, Matt Drobnik, Scott, Naval S.\ under CC-BY.  }
    }
    }
  \label{fig:teaser}
  \end{center}
\end{teaserfigure}

%% file: sections/01_intro.tex
\section{Introduction}
\label{sec:introduction}

Reconstructing a 3D mesh from a collection of 2D images is a long-standing goal in computer vision and graphics. 
Neural field--based 3D reconstruction methods (\emph{e.g.}, \cite{oechsle2021unisurf, wang2021neus, yariv2021volume}) have recently gained 
traction as they can 
reconstruct high-fidelity meshes for both objects and scenes, surpassing the quality of traditional reconstruction pipelines~\cite{schonberger2016pixelwise,openmvs2020}.
But these methods are often demonstrated in controlled capture settings. Consider the Internet photo collections of the Brandenburg Gate and Trevi Fountain shown in \figref{fig:teaser}. Can neural 3D reconstruction techniques apply to real-world, unconstrained Internet datasets like these? 
Handling such data requires both scalability and robustness to highly diverse imagery.

Unlike standard 3D reconstruction datasets that typically contain tens of images (e.g., the DTU dataset~\cite{jensen2014large} provides 49 or 64 images per scene), Internet datasets can contain hundreds or thousands of images.
Neural 3D reconstruction techniques must process such image collections efficiently, without sacrificing accuracy in complex scenes featuring geometric detail of varying granularity.
Beyond scalability to large scenes and image collections, existing reconstruction methods typically assume constant illumination and leverage photometric consistency across the input images. In contrast, for in-the-wild scenes, robustness to appearance variation is another key requirement.

In this work, we present an approach that can efficiently reconstruct surface geometry for large-scale scenes in the presence of varying illumination. Inspired by Neural Radiance Fields in the Wild (NeRF-W)~\cite{martin2021nerf}, we model appearance variation using appearance embeddings, but seek meshes as output, rather than radiance fields as in that work. Meshes, unlike raw radiance fields, provide a direct representation of the scene's geometry and can be readily imported into standard graphics pipelines. To reconstruct such surface geometry, we leverage volume rendering methods as in NeuS~\cite{wang2021neus}, coupling a neural surface representation with volumetric rendering.
However, a straightforward integration of the surface representation presented in NeuS with a volumetric radiance field that models appearance variations involves huge compute demands for large-scale Internet collections, and is intractable in settings with limited access to high-end GPUs. 
For each scene we consider, training using this integrated framework on 32 GPUs converges after roughly ten days.

We therefore propose a hybrid voxel- and surface-guided sampling technique. 
We observe that the standard ray sampling strategy for optimizing neural radiance fields is highly redundant (Figure \ref{fig:sampling}a). To reduce redundant training samples, we first leverage the sparse point clouds from structure-from-motion (SfM) to initialize a sparse volume from which samples are generated (Figure \ref{fig:sampling}b). We then combine this voxel-guided strategy with a surface-guided sampling technique which generates samples based on the current state of optimization (Figure \ref{fig:sampling}c). %
\added{Our key insight here is to not only use the SfM point clouds, but also our surface approximation, yielding new samples that are centered around the true surface. This strategy guides the network to explain the rendered color with near-surface samples, leading to more accurate geometric fitting.}

Finally, while established benchmarks and evaluation schemes exist for controlled datasets, such benchmarks with ground truth geometry do not exist for Internet collections. 
Therefore, we introduce \emph{\datasetname}, a new benchmark dataset derived from the public catalog of free-licensed LiDAR data available in Open Heritage 3D, a repository of open 3D cultural heritage assets.\footnote{ \url{https://openheritage3d.org/}} 
We pair this unique data source with Internet-derived image collections and SfM models from the MegaDepth dataset~\cite{MegaDepthLi18}, performing additional processing steps such as model alignment and visibility checking. 
We also carefully design an evaluation protocol suited for such large-scale scenes with incomplete ground truth (as even LiDAR scans may not cover the entirety of a scene visible from imagery).
Evaluating on \datasetname, we demonstrate that our approach surpasses classical and neural reconstruction methods in terms of efficiency and accuracy.

%% file: sections/02_related_work.tex
\section{Related Work}
\input{figures/sampling.tex}

Image-based 3D reconstruction aims at estimating the most likely 3D shape (and possibly appearance) of an object or a scene given a set of captured 2D photos.
In this section, we summarize prior work ranging from classical to modern approaches, highlighting work most closely related to our own.

\medskip
\paragraph{Multi-view reconstruction}
Multi-view 3D reconstruction methods take images and estimate geometry 
with a variety of representations, including point clouds, depth maps, meshes, or volumetric implicit functions~\cite{Furukawa2015MultiViewSA}. Many classical multi-view stereo (MVS) methods reconstruct geometry by estimating a depth map for each image followed by depth fusion to obtain dense point clouds~\cite{goesele2006multi, furukawa2009accurate, schonberger2016pixelwise, hedman2017casual}. Surface reconstruction algorithms such as Poisson reconstruction~\cite{kazhdan2006poisson, kazhdan2013screened} and Delaunay triagulation~\cite{labatut2009robust} can then be applied to these point clouds to produce meshes. 
Recently, learning-based multiview depth estimation methods have achieved state-of-the-art performance on numerous benchmarks by taking advantage of data-driven priors and physical constraints~\cite{huang2018deepmvs, yao2018mvsnet, Liu2019NeuralRS, yao2019recurrent, gu2020cascade, wilddeepmvs}.%
Since these methods perform depth estimation, point cloud fusion and mesh extraction stages separately, they are sensitive to outliers or inconsistencies in depth maps and can yield noisy or incomplete reconstructions.
In contrast, our approach models the full scene using a global representation and optimizes both appearance and geometry end-to-end via neural rendering.
Other methods also directly estimate the 3D surfaces~\cite{murez2020atlas, bozic2021transformerfusion, sun2021neuralrecon}, but require ground-truth 3D reconstructions for training and cannot generalize beyond the training data (e.g., from indoor scenes to outdoor scenes) due to their data-dependent nature.

\medskip
\paragraph{Reconstruction in the wild}
Internet photo collections, especially 
those capturing tourist attractions around the world, are a popular 
data source in 3D vision and graphics. Because of their abundance and diversity of viewpoints, appearance, and geometry, 
prior research has used such data in a range of problems, including SfM~\cite{snavely2006photo,agarwal2011building,schonberger2016structure}, MVS~\cite{goesele2007multi,furukawa2010towards, frahm2010building, schonberger2016pixelwise}, time-lapse reconstruction~\cite{matzen2014scene, martin20153d, martin2015time} and appearance modeling~\cite{shan2013visual, kim2016multi, meshry2019neural, li2020crowdsampling, martin2021nerf}.
\added{In particular, NeRF-W~\cite{martin2021nerf} models the scene with neural radiance fields, which are suitable for synthesing novel-view images but cannot produce high-quality surface reconstrucions.
In contrast, our approach models the scene with a surface-based representation and directly produces smooth and accurate 3D meshes.}

\medskip
\paragraph{Neural implicit representations}
Neural implicit representations have recently shown great promise for 3D modeling due to their intrinsic global consistency and continuous nature. These properties allow for efficient representation of scene appearance and geometry with a high degree of detail. 
These representations have been applied to a variety of applications including shape generation and completion~\cite{mescheder2019occupancy, park2019deepsdf, peng2020convolutional, chabra2020deep}, novel view synthesis~\cite{mildenhall2020nerf, park2021hypernerf, wizadwongsa2021nex, li2021neural}, camera pose estimation~\cite{yen2020inerf, lin2021barf}, and intrinsic decomposition~\cite{ boss2021neural, boss2021nerd, zhang2021ners, zhang2021nerfactor}.

Most neural implicit representations are optimized from 2D images using differentiable rendering, and can be roughly categorized into two types: surface rendering (\emph{e.g.}, ~\cite{yariv2020multiview,niemeyer2020differentiable}) and volume rendering (\emph{e.g.}, \cite{mildenhall2020nerf, martin2021nerf, kondo2021vaxnerf}, see \cite{dellaert2020neural} for a comprehensive survey). While surface rendering methods allow for more accurate modeling of geometry, most prior methods require additional constraints, such as ground truth masks.
On the other hand, volume rendering techniques have shown impressive results for image-based rendering of complex scenes, but due to their soft volumetric properties it is hard to extract accurate surface geometry from such representations.
More recently, several methods unify surface and volumetric representation, enabling reconstruction of accurate surfaces without requiring masks~\cite{oechsle2021unisurf, wang2021neus, yariv2021volume}. In our work, we extend the representation proposed in NeuS~\cite{wang2021neus} such that it can accommodate unconstrained Internet photo collections.

In contrast to recent work on landmark- or city-scale neural rendering~\cite{rematas2021urban, xiangli2021citynerf, martin2021nerf}, which mainly addresses novel view synthesis, our work focuses on modeling geometry from unstructured Internet photos. We demonstrate that our approach enables efficient, accurate, and highly detailed surface reconstructions of landmarks.

%% file: figures/sampling.tex
\fboxsep=0pt %
\fboxrule=0.4pt %

\begin{figure*}[ht]
\begin{center}
	\captionsetup[sub]{labelformat=parens}
	\includegraphics[width=.92\linewidth]{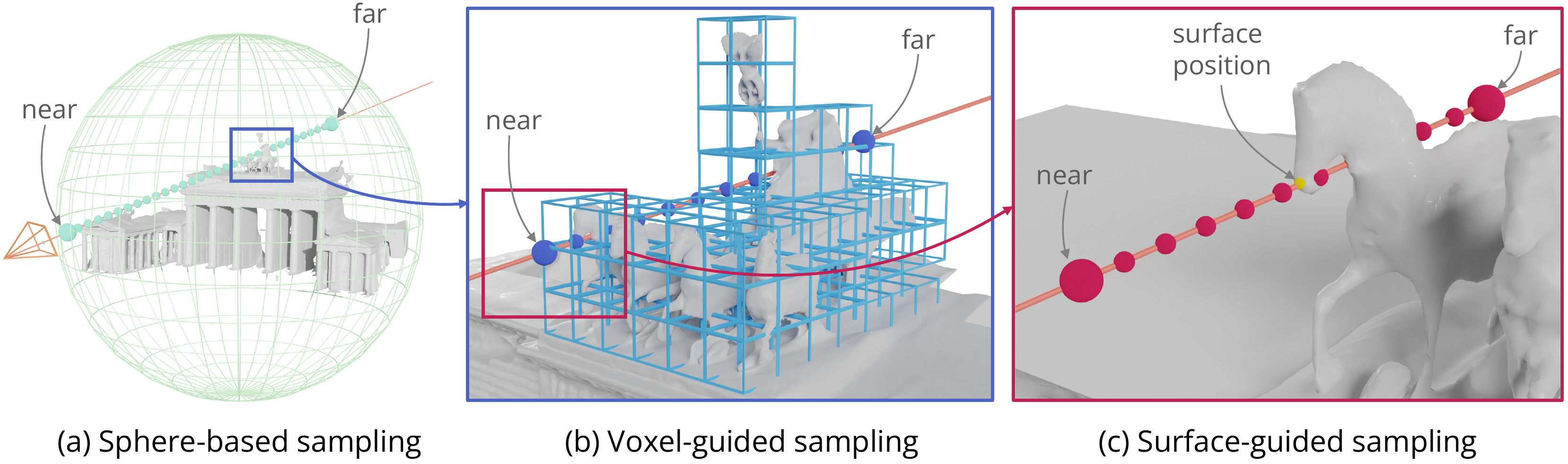}
	\caption{ 
	Comparison between sphere-based sampling and our proposed sampling strategy.
      \normalfont{
    Sphere-based sampling (a), used in NeuS, generates samples scattered throughout the unit sphere and spanning the whole scene, with the result that most samples lie in empty regions and are hence unnecessary.
	We propose voxel-guided sampling (b) to avoid unnecessary samples by sampling only within a sparse voxel volume around surfaces estimated from SfM point clouds (only a subset of voxels are shown for clarity).
	To further increase the sampling density around surfaces, we additionally propose a surface-guided sampling strategy (c), where we store SDF values from previous training iterations in the sparse voxels, and generate samples within a smaller range centered around the estimated surface positions. Note that each successive region of the volume considered by each sampling strategy from (a) to (b) to (c) is progressively smaller as suggested by the 2D blue and red bounding boxes.
	}}
	\label{fig:sampling}
\end{center}
\end{figure*}

%% file: sections/03_method.tex
\section{Method}

To model the shape and appearance of a 3D scene, we propose an approach inspired by recent work on neural radiance fields that can reconstruct a 3D scene as the weights of a neural network by optimizing for image reconstruction losses~\cite{mildenhall2020nerf}. In particular, we use the latent appearance modeling introduced in NeRF in the Wild (NeRF-W)~\cite{martin2021nerf} to model 3D scenes from unconstrained Internet collections with varying lighting. Furthermore, to model accurate surface geometry, we extend the scene representation proposed in NeuS~\cite{wang2021neus} and represent the scene using two neural implicit functions $d$ and $c_i$ encoded by MLPs. Given a point $\mathbf{x}\in \mathbb{R}^3$ in the scene, a viewing direction $\mathbf{v} \in \mathbb{S}^2$ and an image index $i$, we have:
\begin{equation}
d =\mathbf{MLP}_{\mathsf{SDF}}(\mathbf{x}),
\end{equation}
\begin{equation}
c_i = \mathbf{MLP}_{\mathsf{COLOR}}(\mathbf{x},\mathbf{v}, \mathbf{e}_i),     
\end{equation}
where $\{\mathbf{e}_i \}_{i=1}^N$ are appearance embeddings corresponding to each input photo, optimized alongside the parameters of MLPs. 

We use the function $d$ to approximate the signed distance to the true surface $S$, represented as the zero level set of this function:
\begin{equation}
S = \{\mathbf{x}|d(\mathbf{x})=0 \}.
\end{equation}

The function $c_i$ 
models the appearance of 3D point $\mathbf{x}$ as it appears in a given image 
$i$, allowing for the varying appearance of each input image. The MLP parameters and appearance embeddings are learned by optimizing color consistency between real photos and rendered images via a volume rendering scheme: Given a ray, $\{\mathbf{r}(t)=\mathbf{o}+t\mathbf{v}|t\geq 0 \}$ with $\mathbf{o}$ denoting the camera center, 
we can render that ray's expected color $\mathbf{\hat{C}}_i(\mathbf{r})$ corresponding to image $i$ as: 
\begin{equation}
    \mathbf{\hat{C}}_i(\mathbf{r}) = \int_0^{+\infty} w(t)c_i(\mathbf{r}(t),\mathbf{v},\mathbf{e}_i)dt,
\end{equation}
where $w(t)$ is an unbiased and occlusion-aware weight function, as further detailed in \cite{wang2021neus}.

Note that dynamic objects, which are prominent in Internet collections, can significantly impact model performance. The model proposed in NeRF-W~\cite{martin2021nerf}, for instance, incorporates a transient head to distinguish between static and dynamic parts of the scene. To reconstruct accurate geometry, a different approach is required, as the transient head dominates the rendered color, leading to all scene structures modeled as view-dependent transient effects rather than geometry. We discuss our solution and additional design choices in Section \ref{sec:details}. 

\input{figures/compare_sample.tex}

\subsection{Efficient Sampling during Training}
NeuS uses a hierarchical importance sampling strategy to generate sample points on each ray during the optimization phase.
For each scene, NeuS defines a unit bounding sphere to separate the background and foreground parts of the scene. 
The coarse sample points are sampled regularly along a ray between the two intersection points of the ray and the bounding sphere.  Fine-level samples are iteratively generated based on samples from the previous iteration.

This simple strategy works well on lab-captured datasets like DTU, where the camera views are distributed uniformly on a hemisphere.
However, it is extremely inefficient for ``in-the-wild'' scenarios with large-scale scenes, where the camera views are distributed non-uniformly and are often front-facing.
To give an example, if we were to follow NeRF-W, which uses 1024 samples for both the coarse and fine levels per ray for training, 
then using the same number of samples for training the NeuS model would result in an estimated \char`\~10 days of training with 32 GPUs.
Instead, we introduce a hybrid voxel- and surface-guided sampling strategy to improve training efficiency, as detailed in the following sections.
A visualization of different sampling strategies is shown in \figref{fig:sampling}.

\paragraph{Voxel-guided sampling}
To speed up training, we first remove unnecessary training samples by reducing the search space from the entire unit sphere to a smaller spatial envelope that contains the true surface position. Specifically, we observe that rough initial surface estimates are provided by the sparse point cloud that SfM produces alongside the estimated camera poses. Therefore, at the start of training we generate a sparse volume $\mathbf{V}_\mathsf{sfm}$ from the sparse SfM point cloud. This sparse volume is expanded via a 3D dilation operation to ensure that most of the visible surfaces are encompassed by this volume.
The sampling range of a given ray can then be reduced to the in- and out-intersection points between each ray and $\mathbf{V}_\mathsf{sfm}$, and $n_{v}$ points are sampled during this stage. 
We call this sampling technique \emph{voxel-guided sampling}. Related sampling strategies have been explored in recent work~\cite{liu2020neural}, but rather than pruning from a dense voxel grid, our method makes use of the already available 3D information from SfM as more explicit guidance for reducing the search space for point sampling. Moreover, we found that the constructed sparse voxels provides a rough separation of the scene into foreground and background regions, and by removing rays that do not intersect the sparse voxels (e.g., background rays in the sky), the number of required training rays can often be reduced by over 30\%.

\paragraph{Surface-guided sampling}
In order to train the geometry MLP $d$
to accurately fit the 3D surface, it is beneficial to generate as many samples around the true surface as possible.
NeuS achieves a high sampling density through multiple iterations of fine-level importance sampling, which gradually guides samples towards the surface position.
This strategy is time-consuming, since a large number of unnecessary samples must be generated by passing through the geometry MLP $d$ for multiple iterations. 
A visual illustration can be found in \figref{fig:compare_sample}.

Therefore, we propose a surface-guided sampling strategy that further increases sample density around the true surface. In particular, after the training is bootstrapped by voxel-guided sampling, we leverage the surface position estimates from the previous training iteration to generate new samples. 
To achieve this, we cache the SDF predictions from previous iterations inside sparse voxels $\mathbf{V}_\mathsf{cache}$, and query the surface position from this cache at each training iteration.
$\mathbf{V}_\mathsf{cache}$ is an octree built upon $\mathbf{V}_\mathsf{sfm}$ with a depth level of $\ell$.
With the queried surface position $\hat{x}$, we query a number $n_s$ of samples within a narrower range $(\hat{x} - t_s, \hat{x} + t_s)$ around the surface position.
$\mathbf{V}_\mathsf{cache}$ is updated periodically during training to ensure that the stored SDF values are up-to-date.

The cached surface positions provide a good approximation of the true surface position, leading the network to improve upon previous estimations.
\added{Surface-guided sampling 
guides the network to explain the rendered color with samples around the true surface position, allowing the network to fit geometry more accurately.}
As shown in ablation studies, 
without surface-guided sampling, training is unable to converge to the same degree of accuracy even given sufficient time.

\paragraph{Hybrid sampling}
Using only surface-guided sampling will result in artifacts around voxel borders since there is insufficent supervision for empty space.
To avoid this problem, we use a hybrid of voxel- and surface-guided sampling.
Note that the voxel-guided samples are much sparser than the surface-guided samples since they are generated within a much larger search range.
We use another iteration of importance sampling after surface-guided sampling to ensure a good sampling density, bringing the total number of samples along each ray to $n_v + 2 \times n_s$.

\subsection{Additional Details}
\label{sec:details}
\paragraph{Handling transient objects}
We empirically found that, if we directly use the dynamic object modeling strategy proposed in NeRF-W (i.e., a transient NeRF head), the transient NeRF will dominate the rendered color. 
As a result, all scene structures will be modeled as view-dependent transient effects by NeRF instead of by the geometry MLP $d$, since $d$ converges more slowly compared to NeRF. 
We instead use segmentation masks to remove rays belonging to dynamic objects during training.

\paragraph{Supervision signals and handling the textureless sky}
Following NeuS, we use an $\mathcal{L}_1$ loss to supervise the rendered color images ($\mathcal{L}_{\mathsf{COLOR}}$) and use an eikonal term $\mathcal{L}_{\mathsf{REG}}$ to regularize the SDF. 
Since the textureless sky lacks motion parallax, directly using a background NeRF for foreground-background separation as in NeuS will lead to reconstructions contained in spherical shells.
The remaining background rays in $V_\mathsf{sfm}$ (mostly sky) are labeled with semantic masks and penalized as free space with $\mathcal{L}_{\mathsf{MASK}}$. 
Since the semantic masks for the background are not perfect and often contain foreground scene structures, we only apply $\mathcal{L}_{\mathsf{MASK}}$ with a small weight, which we found empirically can remove the sky while keeping foreground structures intact.
Please refer to NeuS for further details on $\mathcal{L}_{\mathsf{COLOR}}$, $\mathcal{L}_{\mathsf{REG}}$ and $\mathcal{L}_{\mathsf{MASK}}$.

\input{figures/gt_alignment.tex}
\section{The \datasetname Benchmark}
To evaluate 
our method we need ground truth 3D geometry. 
However, to the best of our knowledge, there is no existing dataset pairing Internet photo collections with ground truth 3D.
Therefore, 
we introduce \emph{\datasetname}, a new benchmark dataset, derived from Open Heritage 3D. 
We first describe how we constructed the dataset, including how the data was collected and processed (e.g., alignment to the SfM sparse point clouds), and later present the metrics and evaluation protocol used in our experiments.  

\paragraph{Data collection and processing} 
We 
obtained 3D LiDAR data from Open Heritage 3D, which provides public, 
freely-licensed 3D scans 
for hundreds of cultural heritage sites.
We select four landmarks, namely \emph{Brandenburg Gate (BG)}\footnote{\url{https://openheritage3d.org/project.php?id=d51v-fq77}}, \emph{Pantheon Exterior (PE)}\footnote{\url{https://openheritage3d.org/project.php?id=t9sj-mf53}}, \emph{Lincoln Memorial (LM)}\footnote{\url{https://openheritage3d.org/project.php?id=90yg-1054}} and \emph{Palacio de Bellas Artes (PBA)}\footnote{\url{https://openheritage3d.org/project.php?id=vdae-mr89}} for the dataset. 
These landmarks were selected because they can be easily paired with Internet image collections.
The corresponding images for \emph{BG}, \emph{PE} and \emph{LM} were gathered from the MegaDepth dataset~\cite{MegaDepthLi18}.
We collect images for \emph{PBA} from Flickr, 
following a similar procedure as described in prior works for the other landmarks.
The images for each scene are registered using SfM~\cite{schonberger2016structure} to obtain camera poses and a sparse point clouds.
For \emph{BG} and \emph{PBA}, the original LiDAR scans are very dense and are over 100GB in size. 
These point clouds are downsampled to a density of 2\emph{cm}.
A bounding box is manually selected for each scan as the Region of Interest (ROI) to further reduce the size of the point cloud.

\paragraph{Coordinate alignment}
Since the SfM reconstructions and LiDAR scans have different coordinate frames, they must be aligned before evaluation. To align them, we first filter sparse points 
obtained from SfM by their track length and reprojection error, and align the resulting point cloud to the LiDAR scan using ICP~\cite{rusinkiewicz2001efficient} with carefully tuned parameters.
The alignment quality can be visually inspected in \figref{fig:gt_alignment}.
We quantitatively check the alignment quality by reprojecting a set of 
feature tracks from SfM using depth maps rendered from the LiDAR scan. We observe that the resulting reprojection error is less than one pixel across all the scenes, an accuracy level comparable to SfM.

\paragraph{Visibility check}
The LiDAR scans and the images cover different portions of the scene. Only the regions of the scan visible to the input images should be used for evaluation.
We derive visibility information for the LiDAR scans from the SfM point cloud, which is guaranteed to be observable by the images.
To maximize the coverage of the true visible region, we use LoFTR~\cite{sun2021loftr} to run SfM and generate semi-dense point clouds. 
We filter the LiDAR points by generating voxels around the SfM point clouds with a relatively large voxel size.

%% file: figures/compare_sample.tex
\fboxsep=0pt %
\fboxrule=0.4pt %

\begin{figure}[ht]
\begin{center}
	\captionsetup[sub]{labelformat=parens}
	\includegraphics[width=\linewidth]{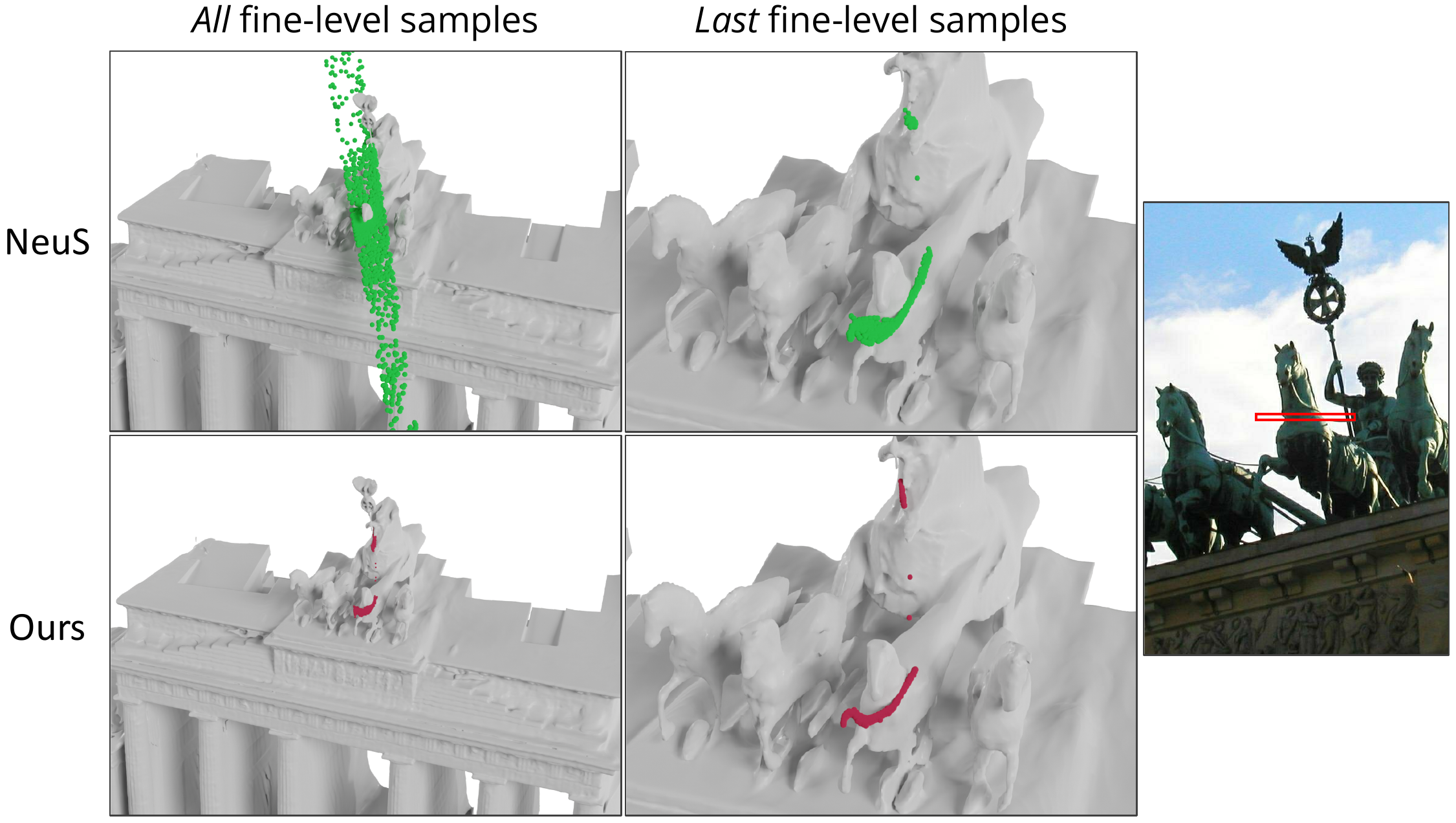}
	\caption{ 
		\textbf{Comparison of fine-level samples from our surface-guided sampling and the sphere-based hierarchical sampling used in NeuS}.
		\normalfont{
		We visualize samples from rays corresponding to pixels in the red box shown in the image on the right.
		The hierarchical sampling in NeuS (upper left) uses a redundant set of fine-level samples (1024 in total), 
		while surface-guided sampling (lower left) uses much fewer samples (24 in total).
		At the last iteration of fine-level sampling, samples from surface-guided sampling (lower right) are denser and closer to the surface than those from NeuS (upper right),
		\added{guiding the network to fit the surface geometry accurately with more details. } 
		The sampled points are generated from trained models.
		The mesh is shown for clarity only.
		}
	}
	\label{fig:compare_sample}
\end{center}
\end{figure}

%% file: figures/gt_alignment.tex
\fboxsep=0pt %
\fboxrule=0.4pt %

\begin{figure}[t]
\begin{center}
	\captionsetup[sub]{labelformat=parens}
	\includegraphics[width=.9\linewidth]{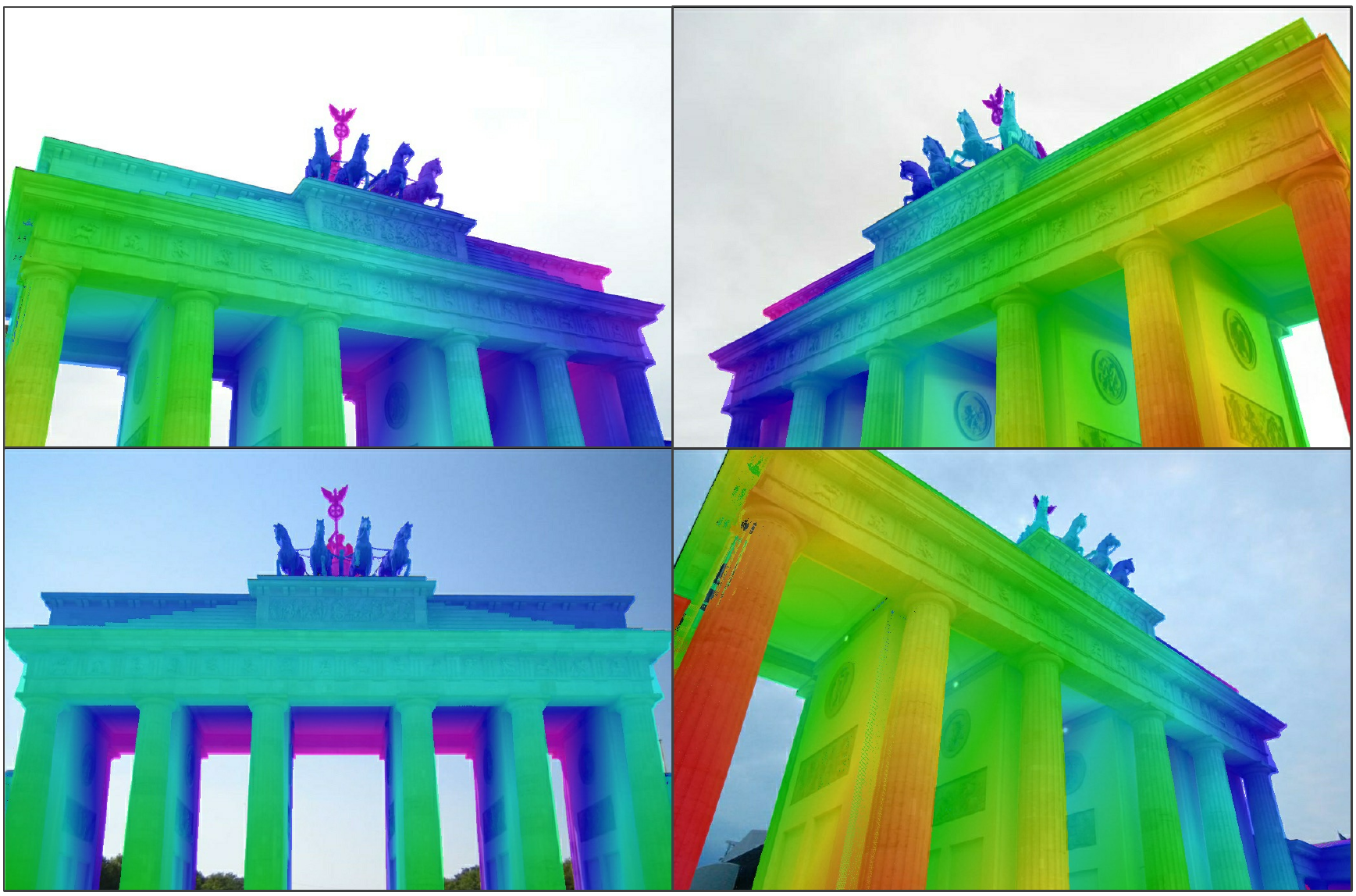}
	\caption{ 
	\textbf{Alignment quality of the LiDAR scan} on \emph{Brandenburg Gate}.
	\normalfont{
	We render the LiDAR-scanned point clouds as depth maps by projecting the points to a set of camera views in the aligned SfM coordinate frame.
	The rendered depth maps are color-coded by depth (warmer colors are closer) and overlaid with the corresponding images. The accuracy of the alignment can be observed in, for instance, the agreement of image and depth edges.
	}}
	\label{fig:gt_alignment}
\end{center}
\end{figure}

%% file: sections/04_experiments.tex
\section{Experiments}

\subsection{Implementation Details}
Training is first bootstrapped by voxel-guided sampling for 5000 iterations, after which  surface-guided sampling is added.
We use 8 layers with 512 hidden units for the geometry MLP and 4 layers with 256 hidden units for the color MLP.
The voxel size $s$ of $\mathbf{V}_\mathsf{sfm}$ for each scene are 2.8, 5.9, 2.0 and 1.0\emph{m} for BG, LM, PE and PBA respectively.
The depth level $\ell$ of octree $\mathbf{V}_\mathsf{cache}$ is 10 for all scenes.
The sampling radius $t_s$ for each scene is defined as $\nicefrac{16}{2^{\ell}}$ times of the voxel size $s$.
We use $n_v = 8$ and $n_s = 8$ in all experiments.
We use 8 NVIDIA A100 GPUs for all the experiments.
For the final output mesh, we only extract a mesh within $\mathbf{V}_\mathsf{sfm}$.

\label{sec:evaluation}

\subsection{Baselines}
We compare our approach to state-of-the-art classical and learning-based MVS algorithms in terms of reconstruction quality and efficiency. For classical methods, we compare against the COLMAP dense reconstruction system~\cite{schonberger2016pixelwise}, which is based on patch-match stereo and Poisson surface reconstruction. We use two different octree depths (11 and 13) in the Poisson reconstruction for comparisons, which we found to have the best numerical accuracy and visual quality, respectively (see \figref{fig:qualitative}). For learning-based approaches, we compare to Vis-MVSNet~\cite{zhang2020visibility}, which achieves state-of-the-art performance on MVS benchmarks. We reconstruct meshes by fusing the depth maps using COLMAP's point fusion algorithm, followed by Poisson surface reconstruction with octree depth=$13$. 
For completeness, we also compare to NeRF-W~\cite{martin2021nerf}. For each scene, we first train a NeRF-W model, then create a pre-defined camera path 
and render color and depth at each viewpoint using the NeRF-W models. We feed the resulting RGB-D sequence into KinectFusion~\cite{izadi2011kinectfusion} from Open3D~\cite{zhou2018open3d} for TSDF fusion to obtain a mesh.

\subsection{Evaluation}

\input{tables/main01_results_}

\paragraph{Metrics and evaluation protocol}
We quantitatively evaluate the reconstruction quality by measuring accuracy and completeness, and evaluate reconstruction efficiency by measuring training/optimization time. Note that compared with object scenarios such as those in DTU, our "in-the-wild" cases exhibit much larger reconstruction errors and thus need multiple thresholds to reflect the reconstruction quality at different scales. 
Therefore, we use three different thresholds (Low, Medium, High) to measure per-scene precision, recall, and F1 scores. 
These thresholds are selected as follows: We first select a maximal threshold $\theta_\mathsf{max}$ to compute the AUC (F1) metric, which integrates the F1 scores from 0 to $\theta_\mathsf{max}$. The maximal threshold is selected as the first threshold that reaches a F1-score of 80 on each scene. The Low, Medium and High thresholds are then evenly sampled between $(0,\theta_\mathsf{max})$.
To better indicate robustness and overall accuracy of each approach, we additionally compute the area under the curve (AUC) of precision, recall and F1 score curves by sweeping the error thresholds. 
More details on the evaluation protocol is presented in the supplemental material.

\input{figures/performance_comp/performance_comp}

\paragraph{Comparisons} The quantitative and qualitative comparisons with baselines are shown in Table~\ref{tab:main_result} and Fig~\ref{fig:qualitative}, respectively. 
For our proposed approach, we report two results: (i) the model after full convergence (ours) and (ii) fast model with early stopping (ours \includegraphics[width=7pt]{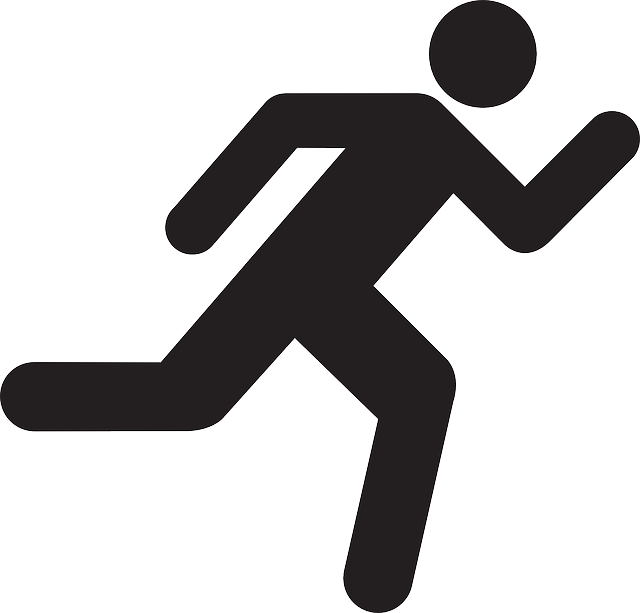}). 
Our approach achieves the best or second best quantitative performance in nearly all scenes and across all thresholds. NeRF-W yields poor reconstruction performance as rendered depth maps are inconsistent across different views (while relatively accurate on individual views). The inconsistent back-projected point clouds will cancel each other’s contribution to the surface position during TSDF fusion, leading to poor reconstruction quality. Even though colmap$^{13}$ outperforms our method on PE and PBA, our visual quality is significantly higher than colmap$^{13}$, as shown in Fig.~\ref{fig:qualitative}. 
In addition, our method achieves the overall best completeness as evidenced by its higher recall scores. 
Our representation and learning objective can even yield filled-in geometry in regions with insufficient observations, while the baselines fail to do so.
To summarize, our method achieves significantly better visual quality than the baselines while being competitive numerically.

\paragraph{Reconstruction time}
We compare the reconstruction time of different methods in~\figref{fig:performance_comp}.
As the figure illustrates, our proposed approach can optimize scenes significantly faster, in most cases also if we consider our models obtained after full convergence (``Ours''). 
The point fusion in COLMAP operates on the CPU and takes the largest portion of the total time, highlighting the advantages of our method as an end-to-end surface reconstruction method.

\input{figures/ablation/ablation}
\paragraph{Ablation studies}
We conduct ablation studies to validate the effectiveness of our proposed  sampling techniques. Fig.~\ref{fig:ablation} shows a comparison of hybrid sampling to sphere-based sampling and pure voxel-guided sampling. Hybrid sampling consistently achieves the best F1 AUC score across almost all time steps, and remains noticeably better than the baselines even after substantial training time.  To achieve an F1 AUC of $\sim$3.2, pure voxel-guided sampling requires 2$\sim$3$\times$ longer training compared to hybrid sampling, and sphere-based sampling is incapable of achieving this accuracy in a reasonable time.  
Additional ablation studies are presented in the supplemental material.

\input{figures/qualitative}

%% file: tables/main01_results_.tex
\definecolor{graytext}{RGB}{130,130,130}

\begin{table}[t]
\setlength{\tabcolsep}{1.0pt}
 \def\arraystretch{0.95}
\centering
\resizebox{\columnwidth}{!}{%
\begin{tabular}{llccclccclccclcccl}
\toprule
\multicolumn{2}{l}{}& \multicolumn{3}{c}{BG} &  & \multicolumn{3}{c}{LM} &  & \multicolumn{3}{c}{PE} &  & \multicolumn{3}{c}{PBA} \\ \cline{3-5} \cline{7-9} \cline{11-13} \cline{15-17} 
& Method %
& P & R & F1 & & P & R & F1 &  & P & R & F1 &  & P & R & F1  \\ \midrule
\multirow{6}{*}{\rotatebox[origin=c]{90}{Low}} & NeRF-W      & 1.8& 1.1& 1.4&  & 28.0& 13.4& 18.1&  & 11.0      & 3.4      & 5.2     &  & 39.6& 12.7& 19.2  
\\
& Vis-MVS       &  48.2& 6.6& 11.6&  & 23.4& 27.0 & 25.1              &  & 36.1     & 22.9      & 28.1    &  & \underline{81.0}& 42.5& 55.7
\\
& colmap$^{11}$       &  61.2& 40.6& 48.8&  & 20.4& 23.6& 21.9&  & \underline{48.7}      & 27.8      & 35.4     &  & \textbf{91.6}& 49.7& 64.5
\\
& colmap$^{13}$       &  \textbf{79.1}& 41.6& \textbf{54.5}&  & 25.8& 32.6& 28.8&  & \textbf{54.4}      & \textbf{71.0}      & \textbf{61.6}     &  & 75.5& 28.9& 41.8
\\
& Ours \includegraphics[width=9pt]{figures/running-icon.png}       &  60.9& \underline{46.8}& 52.9&  & \textbf{33.8}& \textbf{35.9}& \textbf{34.8}&  & 44.7      & 45.9      & 45.3     &  & 77.0& \underline{56.4}& \underline{65.1}
\\
& Ours       & \underline{62.6}& \textbf{47.9}& \underline{54.3}&  & \underline{32.8}& \underline{35.6}& \underline{34.1}&  & 48.5      & \underline{50.8}      & \underline{49.6}     &  & 78.1& \textbf{57.7}& \textbf{66.4}\\ 
\midrule
\multirow{6}{*}{\rotatebox[origin=c]{90}{Medium}} & NeRF-W      & 4.1& 5.0& 4.5&  & 51.9& 24.8& 33.6&  & 21.9      &6.2      & 9.7     &  & 64.3& 28.8& 39.8  
\\
& Vis-MVS       &  72.4& 12.1& 20.8&  & 46.3& 42.7& 44.4&  & 57.8      & 34.4      & 43.1     &  & \underline{95.1}& 57.8& 71.9
\\
& colmap$^{11}$       &  73.7& 53.1& 61.8&  & 50.1& 55.1& 52.4&  & 66.6      & 56.5      & 61.2     &  & \textbf{98.3}& 57.9& 72.9
\\
& colmap$^{13}$       &  \textbf{89.3}& 53.1& 66.6&  & 57.7& 60.6& 59.1&  & \textbf{74.8}      & \textbf{79.7}      & \textbf{77.1}     &  & 86.8& \textbf{71.5}& \textbf{78.4}
\\
& Ours \includegraphics[width=9pt]{figures/running-icon.png}       &  77.6& \underline{64.9}& \underline{70.7}&  & \underline{65.9}& \underline{65.8}& \underline{65.9}&  & 68.3      & 65.8      & 67.1     &  & 87.9& 68.9& 77.3
\\
& Ours        & \underline{78.7}& \textbf{65.7}& \textbf{71.6}&  & \textbf{67.7}& \textbf{68.7}& \textbf{68.2}&  & \underline{71.7}      & \underline{71.1}      & \underline{71.4}     &  & 88.3& \underline{69.7}& \underline{77.9}\\  
\midrule
\multirow{6}{*}{\rotatebox[origin=c]{90}{High}} & NeRF-W      & 7.6& 11.5& 9.2&  & 67.3& 34.2& 45.3&  & 28.8      & 8.3      & 12.9     &  & 79.5& 40.5& 53.7  
\\
& Vis-MVS       &  85.6& 14.5& 24.8&  & 63.8& 53.0& 57.9&  & 68.2      & 42.3      & 52.2     &  & \underline{98.0}& 64.4& 77.7
\\
& colmap$^{11}$       &  81.4& 59.0& 68.4&  &70.7& 74.0& 72.3&  & 73.8      & 69.4      & 71.5     &  & \textbf{99.2}& 63.7& 77.6
\\
& colmap$^{13}$       &  \textbf{94.1}& 58.8&72.4&  & 78.0& 75.9& 76.9&  & \textbf{81.6}      & \textbf{83.6}      & \textbf{82.6}     &  & 91.3& \textbf{80.8}& \textbf{85.7}
\\
& Ours \includegraphics[width=9pt]{figures/running-icon.png}     &  85.3& \underline{73.2}& \underline{78.8}&  & \underline{80.8}&\underline{80.2}& \underline{80.5}&  & 77.1      & 73.2      & 75.1     &  & 92.8& 76.6& 83.9
\\
& Ours       & \underline{85.9}& \textbf{73.7}& \textbf{79.3}&  & \textbf{82.1}& \textbf{82.1}& \textbf{82.1}&  & \underline{79.9}      & \underline{77.9}      &\underline{78.9}     &  & 93.0& \underline{77.1}& \underline{84.3}\\ \midrule
\multirow{6}{*}{\rotatebox[origin=c]{90}{All (AUC)}} & NeRF-W      & 2.0& 2.8& 2.3&  & 2.9& 1.5& 2.0&  & 6.1      & 1.9      & 2.9     &  & 46.4& 21.4& 29.1  
\\
& Vis-MVS       &  26.3& 4.2& 7.2&  & 2.7& 2.4& 2.5&  & 12.4      & 8.3      & 10.0     &  & \underline{66.6}& 40.7& 50.4
\\
& colmap$^{11}$       &  27.4& 18.8& 22.1&  & 2.8& 3.0& 2.9&  & 13.5      & 12.4      & 12.9     &  & \textbf{71.9}& 43.3& 54.0
\\
& colmap$^{13}$       &  \textbf{33.3}& 19.1& 24.1&  & 3.2& 3.2& 3.2&  & \textbf{14.6}      & \textbf{15.7}      & \textbf{14.9}     &  & 62.9& 45.5& 50.5
\\
& Ours \includegraphics[width=9pt]{figures/running-icon.png}      &  28.2& \underline{22.9}& \underline{25.1}&  & \underline{3.3}& \underline{3.3}& \underline{3.3}&  & 13.7      & 13.3      & 13.5     &  & 63.9& \underline{50.4}& \underline{56.3}
\\
& Ours      &  \underline{28.6}& \textbf{23.2}& \textbf{25.5}&  &\textbf{3.4}& \textbf{3.4}& \textbf{3.4}&  &  \underline{14.1}      &  \underline{14.0}      & \underline{14.0}     &  & 64.4& \textbf{51.1}& \textbf{56.9}\\
       \bottomrule 
\end{tabular}
}
\vspace{0.8em}
\caption{\textbf{Reconstruction Evaluation}. \normalfont{We report precision (P), recall (R) and F1 scores over the different datasets in \datasetname. 
We compare our reconstruction performance to NeRF-W~\cite{martin2021nerf}, Vis-MVS~\cite{zhang2020visibility} and two variants of COLMAP, as detailed in the text. 
For our method, ``Ours'' measures performance after training converges and ``Ours \protect\includegraphics[width=7pt]{figures/running-icon.png}'' is an earlier checkpoint, 
selected using the F1 score (taking the model that yields $95\%$ of the final value). We report performance over three different thresholds (``Low'', ``Medium'', ``High'') and also report an AUC metric that integrates performance over all thresholds (``All (AUC)''). Best results are in bold, and second best are underlined. As illustrated above, our method obtains the best or second best performance in nearly all cases.  
}}
\label{tab:main_result}
\end{table}

%% file: figures/performance_comp/performance_comp.tex
\begin{figure}
    \centering
    \includegraphics[width=0.95\columnwidth]{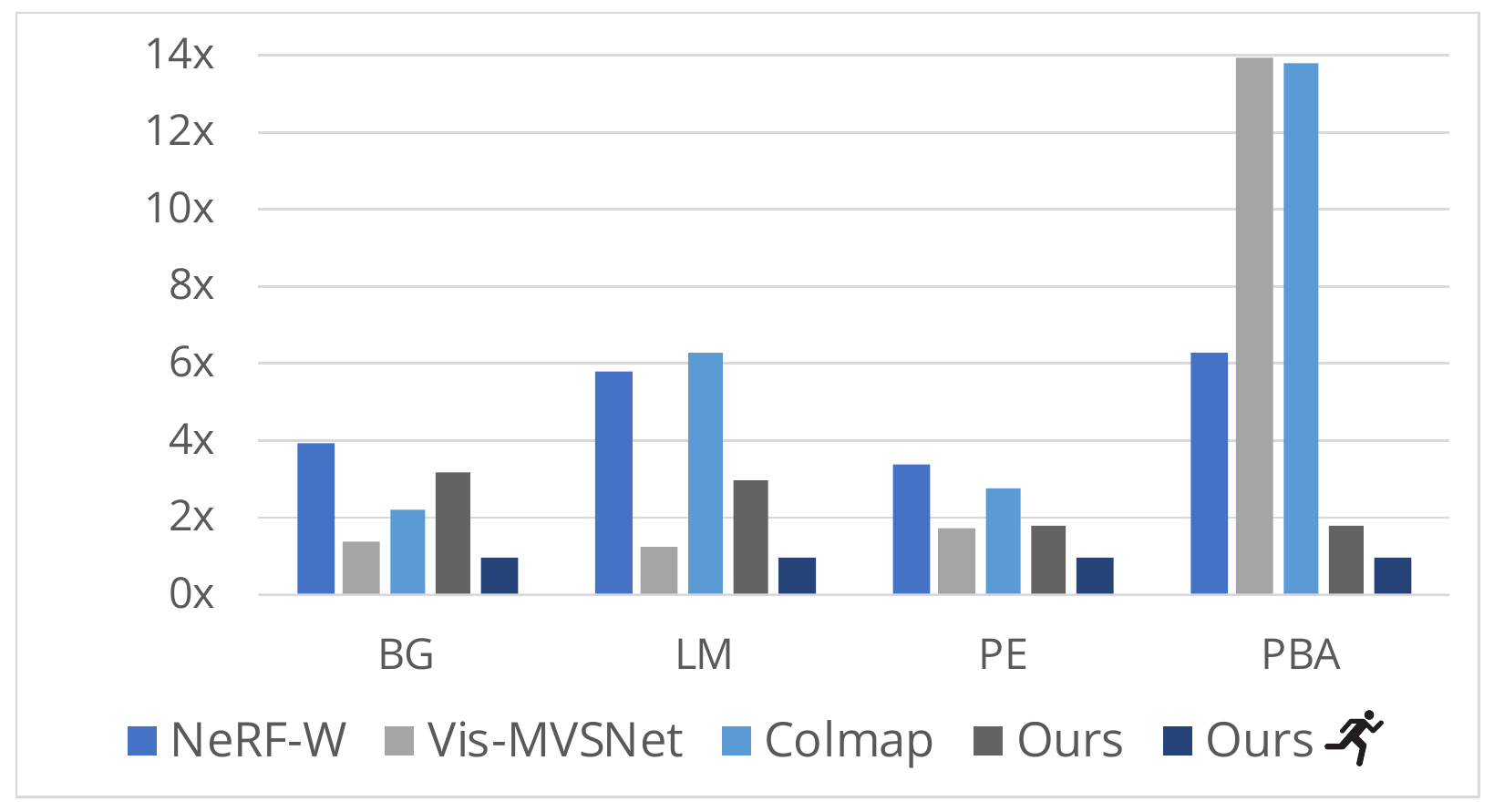}
    \caption{\textbf{Training speed in comparison to ``Ours \protect\includegraphics[width=7pt]{figures/running-icon.png}''}. \normalfont{As illustrated above, for larger scenes (e.g., PBA), the baselines can be up to $14\times$ slower compared to our model that is optimized for training speed. In contrast, our best model yields a slowdown of up to $3\times$.} 
    }
    \label{fig:performance_comp}
\end{figure}

%% file: figures/ablation/ablation.tex
\begin{figure}
    \centering
    \includegraphics[width=\columnwidth]{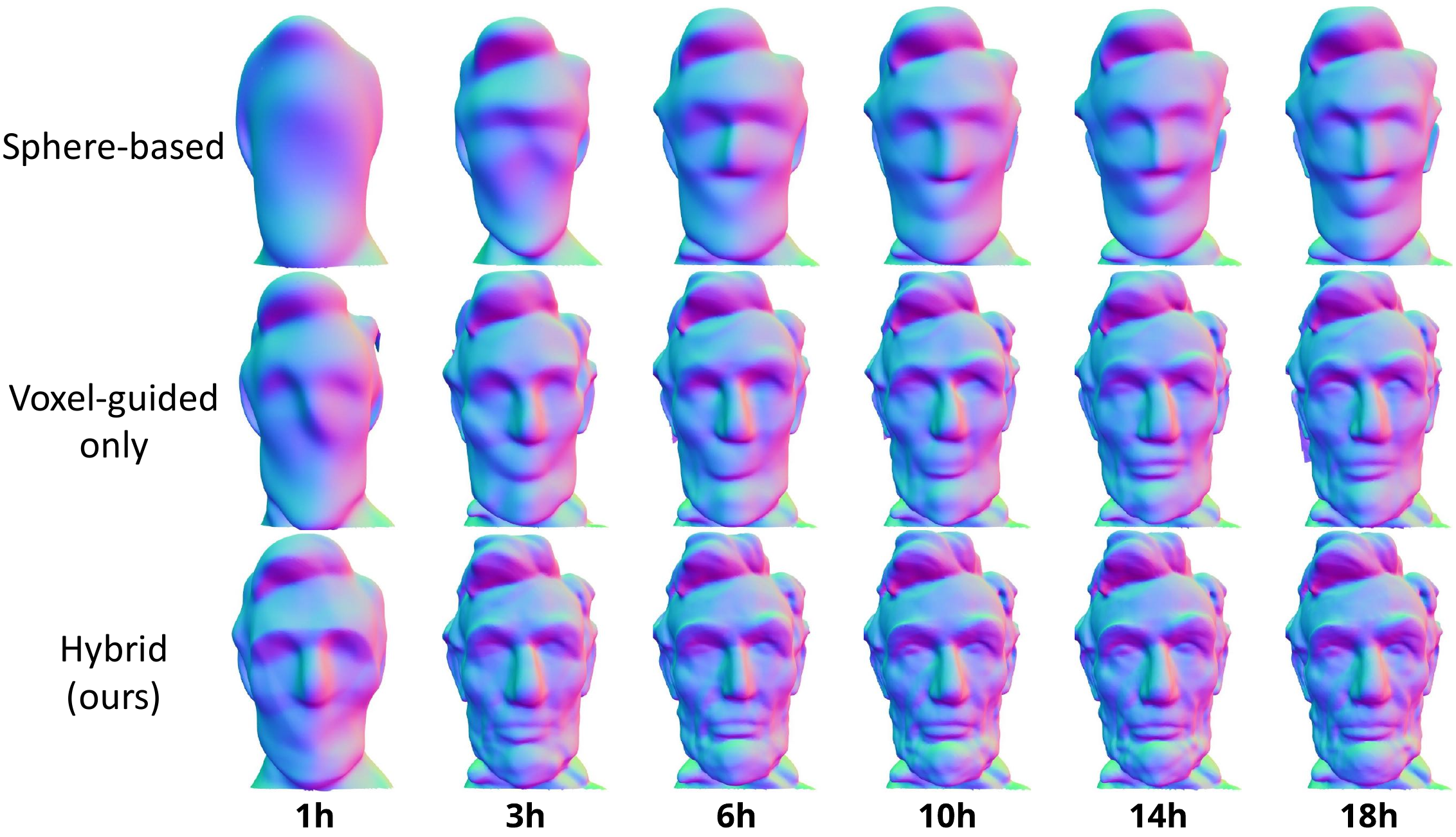}
    \includegraphics[width=\columnwidth]{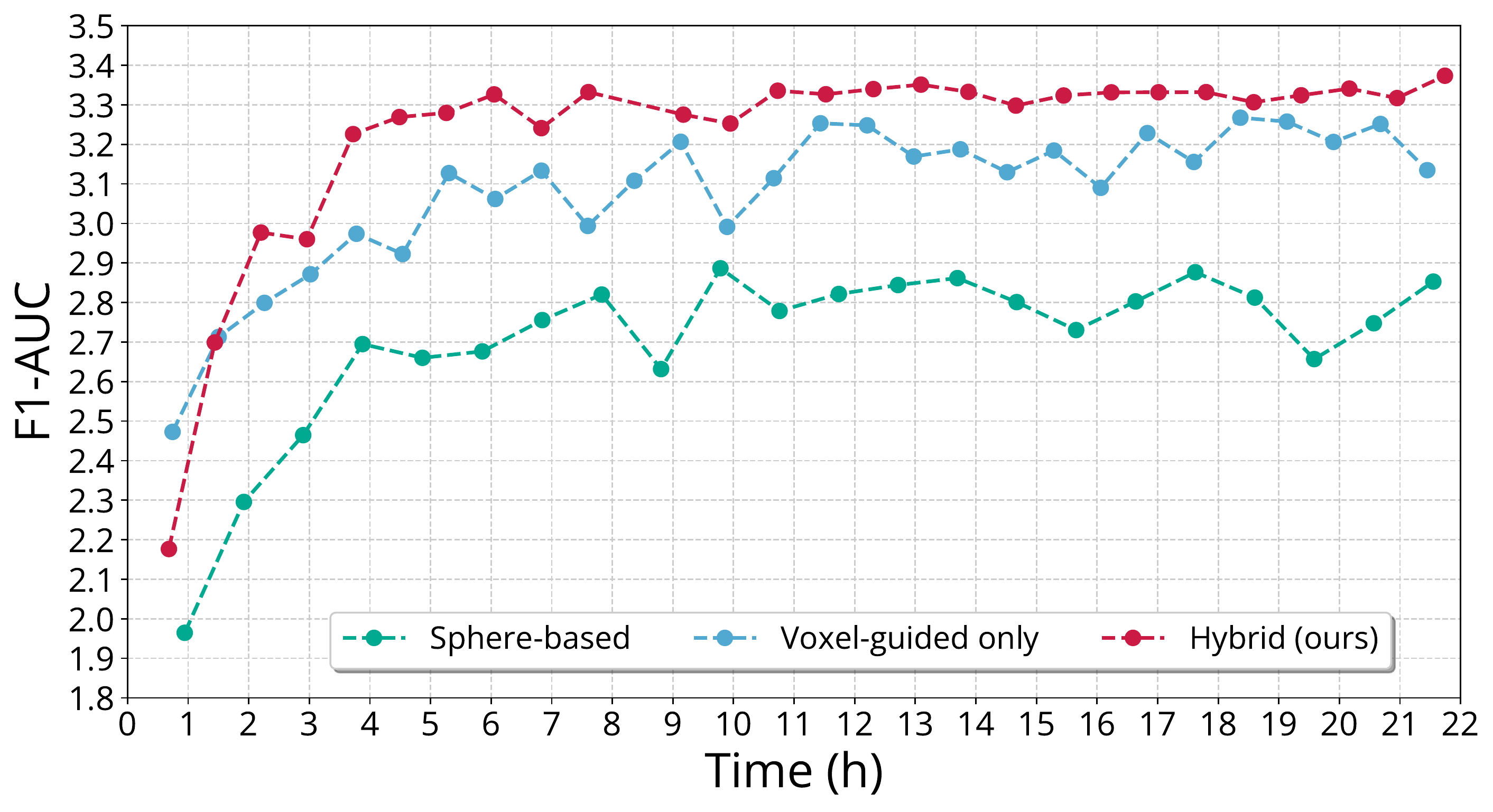}
    \caption{
    \textbf{Ablation studies.} 
    \normalfont{
    We visualize the reconstructed meshes (zoomed in to the head region) and plot the curves of performance w.r.t.\ training time for sphere-based sampling, voxel-guided sampling and hybrid sampling on the Lincoln Memorial (LM). 
    Hybrid sampling leads to significantly better training speed while consistently achieving the best F1-AUC accuracy and visual quality.
    }}
    \label{fig:ablation}
\end{figure}

%% file: figures/qualitative.tex
\fboxsep=0pt %
\fboxrule=0.4pt %

\begin{figure*}
  \begin{center}
    \captionsetup[sub]{labelformat=parens}
    \includegraphics[width=\textwidth]{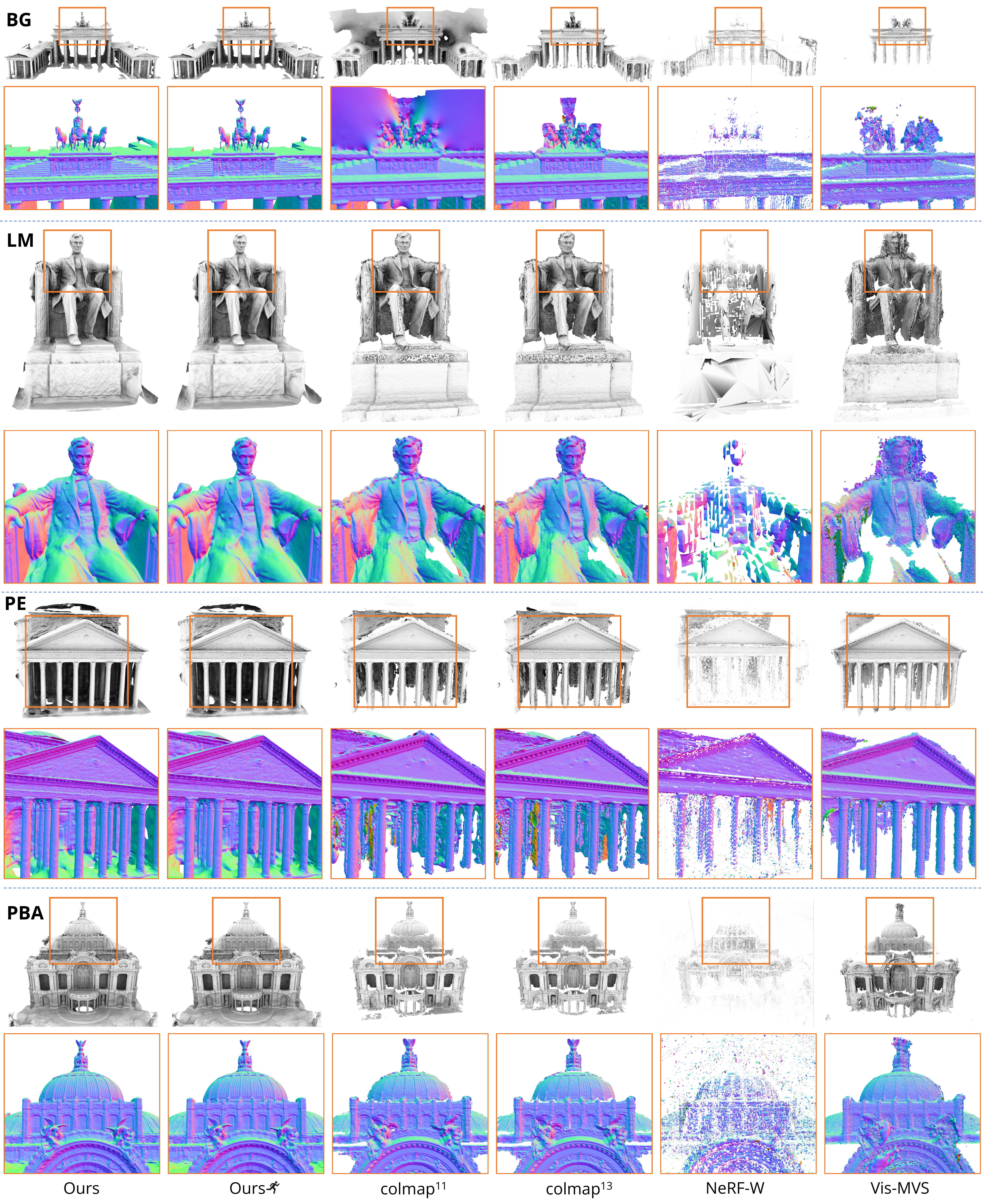}
    \caption{\textbf{Qualitative results on the \datasetname dataset.} \normalfont{As illustrated in the full rendered meshes and the zoomed-in selected regions (in orange), our models yield more complete and higher quality meshes. Zoom in for details. Interactive visualizations can be found at the \href{https://zju3dv.github.io/neuralrecon-w/}{project page}.}}
  \label{fig:qualitative}
  \end{center}
\end{figure*}

%% file: sections/05_conclusion.tex
\section{Limitations and Conclusion}
\paragraph{Limitations} Our approach inherits limitations from NeRF-like methods. For example, inaccurate camera registration can affect final reconstruction quality. In addition, since our model only learns surface locations from known image observations without imposing domain-specific priors, it can fail to produce accurate geometry 
in unseen regions.

\medskip
\paragraph{Conclusion} We presented a new neural %
method for high-quality 3D surface reconstruction from Internet photo collections. To efficiently learn accurate surface locations of complex scenes, we introduce a hybrid voxel-surface guided sampling technique that significantly improves training time over baseline methods. In the future, we envision a full inverse rendering approach,
as well as the ability to model scene dynamics across different time scales.